\documentclass[sigconf, authorversion]{acmart}

\usepackage{booktabs} % For formal tables
\usepackage{tikz}
\usetikzlibrary{arrows,positioning}
\usetikzlibrary{shapes.geometric}
\usepackage{times}
\usepackage{graphicx}
\usepackage{amsmath}
\usepackage{amssymb}
\usepackage[caption=false,font=footnotesize]{subfig}
\usepackage{multirow}
\usepackage[export]{adjustbox}
\usepackage{xcolor}
\usepackage{arydshln}
\usepackage[OT1]{fontenc} 
\usepackage{silence}
\setcopyright{none}
\acmDOI{}

\acmISBN{}

\acmConference[-]{-}{2018}{-}
\acmYear{2018}
\copyrightyear{2018}

\renewcommand\footnotetextcopyrightpermission[1]{}
\begin{document}

\title{Light-weight pixel context encoders for image inpainting}

\author{Nanne van Noord}
\authornote{This work was done while Nanne van Noord worked at Tilburg 
University.}
\affiliation{%
  \institution{University of Amsterdam}
}
\email{n.j.e.vannoord@uva.nl}

\author{Eric Postma}
\affiliation{%
  \institution{Tilburg University}
}
\email{e.o.postma@tilburguniversity.edu}

\begin{abstract}
In this work we propose Pixel Content Encoders (PCE), a light-weight image 
inpainting model, capable of generating novel content for large missing 
regions in images. Unlike previously presented convolutional neural network based models, 
our PCE model has an order of magnitude fewer trainable parameters. Moreover, 
by incorporating dilated convolutions we are able to preserve fine grained spatial 
information, achieving state-of-the-art performance on benchmark datasets of 
natural images and paintings.
Besides image inpainting, we show that without changing the architecture, PCE 
can be used for image extrapolation, generating novel content beyond existing 
image boundaries.
\end{abstract}

\maketitle
\section{Introduction}

\begin{figure}[!ht]
  \centering
  \includegraphics[width=.45\linewidth]{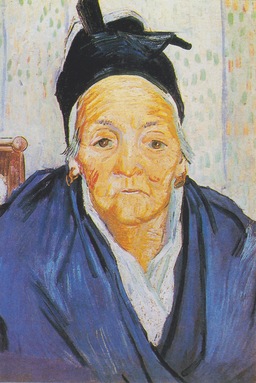}
  \caption{``\textit{An old woman of Arles}'' by Vincent van Gogh (courtesy of 
  the Van Gogh Museum). The left bottom corner was manually inpainted with a 
`neutral' colour.} \label{fig:inpaintexample}
\end{figure}

Reconstructing missing or damaged regions of paintings has long required a 
skilled conservator or artist. Retouching or inpainting is typically only done 
for small regions, for instance to hide small defects \cite{Bertalmio2000}.  
Inpainting a larger region requires connoisseurship and imagination: the 
context provides clues as to how the missing region might have looked, but 
generally there is no definitive evidence.  Therefore, sometimes the choice is 
made to inpaint in a conservative manner.  Take for example the painting in 
Figure~\ref{fig:inpaintexample}, the left bottom corner was filled with a 
`neutral' colour as to not change the interpretation of the artwork.  However, 
with the emergence of powerful computer vision methods specialising in 
inpainting \cite{Chan2011, Cornelis2013, Pathak, Iizuka2017}, it has become 
possible to explore what a potential inpainting result might look like, without 
physically changing the painting. 

Although image inpainting algorithms are not a novel development 
\cite{Bertalmio2000, Barnes2009}, recent work has shown that approaches based 
on Convolutional Neural Networks (CNN) are capable of inpainting large missing 
image regions in a manner which is consistent with the context \cite{Pathak, 
Yeh2016, yang2017,Iizuka2017}.  Unlike, scene-completion approaches 
\cite{Hays2007}, which search for similar patches in a large database of 
images, CNN-based approaches are capable of \textit{generating} meaningful 
content \cite{Pathak}.  

\begin{figure}[!ht]
  \centering
        \begin{tikzpicture}[every node/.style={rectangle, minimum width=0cm, 
        minimum height=1cm, line width=1pt, draw=blue!80},on grid] 
        
        \begin{scope}[xshift=-100]
          \node[minimum width=3cm] (A) at (0,0) {Input};
          \node[minimum height=0.7cm, trapezium, trapezium angle=-60] (B) at 
          (0,-1.25) {Encoder};
          \node[minimum width=2cm, minimum height=0.5cm] (C) at (0,-2.25) 
        {Bottleneck};
          \node[minimum width=2.5cm] (D) at (0,-3.25) {Classifier};
          \node[minimum width=1cm] (E) at (0,-4.5) {Output};
        \draw [->, thick] (A) -- (B);
        \draw [->, thick] (B) -- (C);
        \draw [->, thick] (C) -- (D);
        \draw [->, thick] (D) -- (E);
        \node[above of=A, draw=none] (2) {\textbf{Classifier}};
        \end{scope}

        \begin{scope}[xshift=0]
           \node[minimum width=3cm] (A) at (0,0) {Input};
          \node[minimum height=0.7cm, trapezium, trapezium angle=-60] (B) at 
          (0,-1.25) {Encoder};
          \node[minimum width=2cm, minimum height=0.5cm] (C) at (0,-2.25) 
        {Bottleneck};
          \node[minimum height=0.7cm, trapezium, trapezium angle=60] (D) at 
          (0,-3.25) {Decoder};
          \node[minimum width=3cm] (E) at (0,-4.5) {Output};
        \draw [->, thick] (A) -- (B);
        \draw [->, thick] (B) -- (C);
        \draw [->, thick] (C) -- (D);
        \draw [->, thick] (D) -- (E);
        \node[above of=A, draw=none] (2) {\textbf{Encoder-Decoder}};
        \end{scope}
    \end{tikzpicture}
      \caption{Visualisation of a classification CNN architecture (left), and 
      an image generation architecture (right). In both architectures the 
      encoder downsamples the input into a low(er) dimensional representation: 
      \textit{the bottleneck}.}
      \label{inp:fig:bottleneck}
\end{figure}
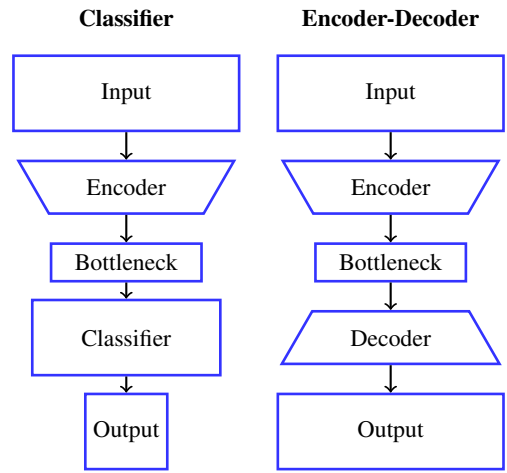

A key aspect of CNN-based inpainting approaches and of many CNN architectures 
in general \cite{Sermanet2011}, is that an image is described at multiple 
scales by an encoder that reduces the spatial resolution through pooling and 
downsampling.  Each layer (or block of layers) of the network processes the 
image at a certain scale, and passes this scale-specific information on to the 
next layer.  This encoding process continues until a single low dimensional 
representation of the image is found, which describes the entire image.  
Because this architecture resembles a funnel, the final representation is 
sometimes referred to as \textit{the bottleneck}.  
Figure~\ref{inp:fig:bottleneck} shows a visualisation of two CNN architectures; 
one for classification, and one for image generation (similar to an 
autoencoder).  Both architectures encode the image into a bottleneck 
representation, after which the classification network processes it with a 
classifier, typically a softmax regression layer \cite{Krizhevsky2012}, and the 
image generation network feeds it to a decoder \cite{Isola2016}. The decoder 
subsequently performs a number of upsampling steps to generate the output 
image.  

A downside of downsampling in the encoder is the loss of spatial detail - 
detail which might be crucial to the task \cite{Yu2017}. For inpainting this is 
most prevalent when considering the local consistency \cite{Iizuka2017}; the 
consistency between the edge of the inpainted region and the edge of the 
context.  A lack of local consistency will result in an obvious transition from 
the context to the inpainted region. Although increasing the size of the 
bottleneck, i.e., making it wider, appears to alleviate this to some extent 
\cite{Pathak}, it comes at the cost of a tremendous increase in model 
parameters. Luckily, recent work has shown that it is possible to encode an 
image while preserving the spatial resolution \cite{Yu2016, Yu2017}.  
\textit{Dilated convolutions} make it possible to expand the receptive field of 
a CNN, without downsampling or increasing the number of model parameters.  We 
define the receptive field of a CNN as the size of the region in the input 
space that affect the output neurons of the encoder. For instance, a single 
layer CNN with $3 \times 3$ filters would have a receptive field of $3 \times 
3$, adding identical layers on top would increase the receptive field to $5 
\times 5$, $7 \times 7$, etc.  We refer to Section~\ref{inp:subsec:dilated} for 
an explanation of how the receptive field of a CNN grows when using dilated 
convolutions.

Many of the shown results obtained with CNN-based inpainting models, have 
been achieved using complex architectures with many parameters, resulting in a 
necessity of large amounts of data, and often long training times \cite{Pathak, Iizuka2017}.
Although simpler architectures have been proposed \cite{Yeh2016}, these are typically only 
demonstrated on small datasets with relatively little variation (i.e., only 
faces or only facades of buildings). Therefore, we aim to produce a light-weight inpainting
model, which can be applied to large and complex datasets.
In this paper, we demonstrate that using dilated convolutions we can construct
a simple model that is able to obtain state-of-the-art performance on various inpainting tasks. 

The remainder of this paper is organised as follows. In 
Section~\ref{inp:sec:relwork} we discuss related work on inpainting and dilated 
convolutions. In \ref{inp:sec:pce} we describe the architecture of our model 
and how it is trained. Section~\ref{inp:sec:exp} describes the experiments and 
the results we obtain on a variety of inpainting tasks. Lastly, in 
Section~\ref{inp:sec:conclusion} we conclude that our model is much less 
complex than existing models, while outperforming them on benchmark datasets of 
natural images and paintings.

\section{Related work}
\label{inp:sec:relwork}

In this section we will discuss work related to image inpainting, dilated 
convolutions and their application to inpainting, and finally our 
contributions. 

\subsection{Image inpainting}

When a single pixel is missing from an image we can look at the adjacent pixels 
and average their colour values to produce a reasonable reconstruction of the 
missing pixel.  When a larger region formed by directly adjacent pixels is 
missing, it is necessary to take into account a larger neighbourhood 
surrounding the missing region. Moreover, it may become insufficient to only 
smooth out the colour, to reconstruct the region in a plausible manner.  
Additionally, for smaller regions it can be sufficient to only incorporate 
textural or structural information \cite{Bertalmio2003}, however inpainting 
larger regions requires understanding of the entire scene \cite{Pathak}.  For 
example, given a picture of a face, if part of the nose is missing it can be 
reconstructed by looking at the local context and textures. But once the entire 
nose is missing it requires understanding of the entire face to be able to 
reconstruct the nose, rather than smooth skin \cite{Yeh2016}.  

The challenge of inferring (large) missing parts of images is at the core of 
image inpainting, the process of reconstructing missing or damaged parts of 
images \cite{Bertalmio2000}.

Classical inpainting approaches typically focus on using the local context to 
reconstruct smaller regions, in this paper we will focus on recent work using 
Convolutional Neural Networks (CNN) to encode the information in the entire and 
inpaint large regions \cite{Pathak, Yeh2016, yang2017, Gao2017, Iizuka2017}.  
From these recent works we will focus on two works, first the work by Pathak et 
al.  \cite{Pathak} who designed the (until now) `canonical' way of performing 
inpainting with CNN.  Second, we will focus on the work by Iizuka et al.  
\cite{Iizuka2017}, who very recently proposed several extensions of the work by 
Pathak et al., including incorporating dilated convolutions. 

Pathak et al. \cite{Pathak} present Context Encoders (CEs), a CNN trained to 
inpaint while conditioned on the context of the missing region. CE describe the 
context of the missing region by encoding the entire image into a bottleneck 
representation. Specifically, the spatial resolution of the input image is 
reduced with a factor $128$; from $128 \times 128$, to the bottleneck 
representation - a single vector. To compensate for the loss of spatial 
resolution they increase the width of the bottleneck to be $4000$ dimensional.  
Notably, this increases the total number of model parameters tremendously, as 
compared to a narrower bottleneck.  

CEs are trained by means of a reconstruction loss (L2), and an adversarial 
loss. The adversarial loss is based on Generative Adversarial Networks (GAN) 
\cite{Goodfellow2014}, which involves training a discriminator $D$ to 
distinguish between real and fake examples. The real examples are samples from
the data $x$, whereas the fake examples are produced by the generator $G$.
Formally the GAN loss is defined as:
\begin{equation}
  \min_{G} \max_{D} \mathbb{E}_{x \sim p_{data}(x)}[log(D(x) + log(1 - D(G(x))]
\end{equation}
by minimising this loss the generator can be optimised to produce examples 
which are indistinguishable from real examples.  In \cite{Pathak} the generator 
is defined as the CE, and the discriminator is a CNN trained to distinguish 
original images from inpainted images.

In a more recent paper, Iizuka et al. \cite{Iizuka2017} propose two extensions 
to the work by Pathak et al. \cite{Pathak}: (1) They reduce the amount of 
downsampling by incorporating dilated convolutions, and only downsample by a 
factor $4$, in contrast to Pathak et al. who downsample by a factor $128$.  (2) 
They argue that in order to obtain globally and locally coherent inpainting 
results, it is necessary to extend the GAN framework used in \cite{Pathak} by 
using two discriminators. A `local' discriminator which focuses on a small 
region centred around the inpainted region, and a `global' discriminator which 
is trained on the entire image.  Although the qualitative results presented in 
\cite{Iizuka2017} appear intuitive and convincing, the introduction of a second 
discriminator results in a large increase in the number of trainable 
parameters. 

Ablation studies presented in a number of works on inpainting have shown that 
the structural (e.g., L1 or L2) loss results in blurry images 
\cite{Pathak,yang2017,Iizuka2017}. Nonetheless, these blurry images do 
accurately capture the coarse structure, i.e., the low spatial frequencies.  
This matches an observation by Isola et al. \cite{Isola2016}, who stated that 
if the structural loss captures the low spatial frequencies, the GAN loss can 
be tailored to focus on the high spatial frequencies (the details).  
Specifically, Isola et al. introduced PatchGAN, a GAN which focuses on the 
structure in local patches, relying on the structural loss to ensure 
correctness of the global structure.  PatchGAN, produces a judgement for $N 
\times N$ patches, where $N$ can be much smaller than the whole image.  When 
$N$ is smaller than the image, PatchGAN is applied convolutionally and the 
judgements are averaged to produce a single outcome. 

Because the PatchGAN operates on patches it has to downsample less, reducing 
the number of parameters as compared to typical GAN architectures, this fits 
well with our aim to produce a light-weight inpainting model. Therefore, in our 
work we choose to use the PatchGAN for all experiments.

Before turning to explanation of the complete model in 
section~\ref{inp:sec:pce}, we first describe dilated convolutions in more 
detail.

\subsection{Dilated convolutions}
\label{inp:subsec:dilated}

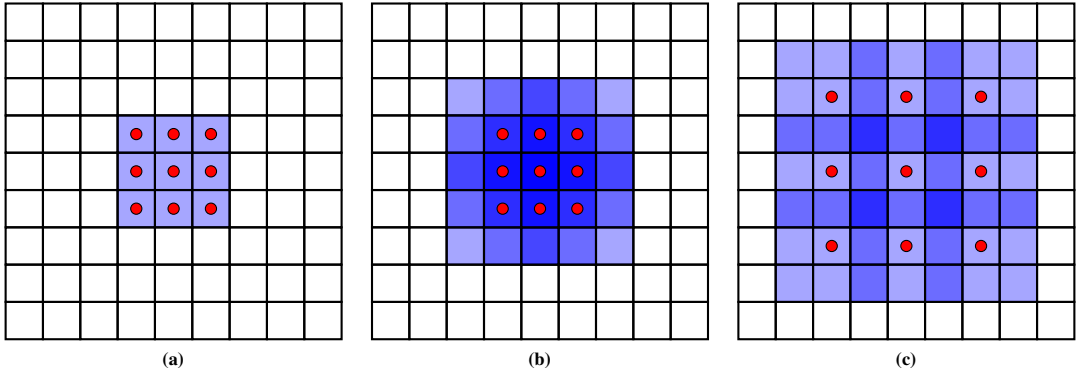
\begin{figure*}[!t]
  \centering
        \resizebox{0.8\textwidth}{!}{%
      \subfloat[]{%
        \begin{tikzpicture}[ box/.style={rectangle,draw=black,thick, minimum 
          size=0.5cm}, dot/.style={draw,shape=circle,fill=red,minimum 
          size=0.15cm,inner sep=0pt}]
          \foreach \x in {0.5,1.0,...,4.5}{
              \foreach \y in {0.5,1.0,...,4.5}
                  \node[box] at (\x,\y){};
          }
          \foreach \x in {2,2.5,...,3}{
              \foreach \y in {2,2.5,...,3} {
                \node[box,fill=blue,opacity=0.35] at (\x,\y){};
                \node[dot,fill=red] at (\x,\y){};
              }}
        \end{tikzpicture}}
        \hspace{0.3cm}
        \subfloat[]{%
        \begin{tikzpicture}[ box/.style={rectangle,draw=black,thick, minimum 
          size=0.5cm}, dot/.style={draw,shape=circle,fill=red,minimum 
          size=0.15cm,inner sep=0pt}]
          \foreach \x in {0.5,1.0,...,4.5}{
              \foreach \y in {0.5,1.0,...,4.5}
                  \node[box] at (\x,\y){};
          }
          \foreach [count=\i,
          evaluate=\i as \xmin using (\x-0.5),
          evaluate=\i as \xmax using (\x+0.5)] \x in {2,2.5,...,3}{
              \foreach [count=\i,
          evaluate=\i as \ymin using (\y-0.5),
          evaluate=\i as \ymax using (\y+0.5)]  \y in {2,2.5,...,3}{
                \foreach \j in {\xmin,\x,...,\xmax}{
                    \foreach \k in {\ymin,\y,...,\ymax}
                      \node[box,fill=blue,opacity=0.35] at (\j,\k){};
                }
              }}
              \foreach \x in {2,2.5,...,3}{
              \foreach \y in {2,2.5,...,3}
                \node[dot,fill=red] at (\x,\y){};
              }
        \end{tikzpicture}}
        \hspace{0.3cm}
        \subfloat[]{%
        \begin{tikzpicture}[ box/.style={rectangle,draw=black,thick, minimum 
          size=0.5cm}, dot/.style={draw,shape=circle,fill=red,minimum 
          size=0.15cm,inner sep=0pt}]
          \foreach \x in {0.5,1.0,...,4.5}{
              \foreach \y in {0.5,1.0,...,4.5}
                  \node[box] at (\x,\y){};
          }
          \foreach [count=\i,
          evaluate=\i as \xmin using (\x-0.5),
          evaluate=\i as \xmax using (\x+0.5)] \x in {1.5,2.5,...,3.5}{
              \foreach [count=\i,
          evaluate=\i as \ymin using (\y-0.5),
          evaluate=\i as \ymax using (\y+0.5)]  \y in {1.5,2.5,...,3.5}{
                \foreach \j in {\xmin,\x,...,\xmax}{
                    \foreach \k in {\ymin,\y,...,\ymax}
                      \node[box,fill=blue,opacity=0.35] at (\j,\k){};
                }
                \node[dot,fill=red] at (\x,\y){};
              }}
        
            \end{tikzpicture}}}
            \caption{Comparison of $1$-dilated versus $2$-dilated filter. (a) 
              shows the receptive field of a $3 \times 3$  $1$-dilated filter 
            directly on the input.  (b) shows the $5 \times 5$ receptive field 
          of a $1$-dilated $3 \times 3$ filter applied to (a). (c) shows the $7 
        \times 7$ receptive field of a $2$-dilated $3 \times 3$ filter applied 
      to (a). (c) has a larger receptive field than (b), with the same number 
    of parameters.}
            \label{inp:fig:dilation}
\end{figure*}

The convolutional layers of most CNN architectures use discrete convolutions.
In discrete convolutions a pixel in the output is the sum of the elementwise 
multiplication between the weights in the filter and a region of adjacent 
pixels in the input. Dilated or $l$-dilated convolutions offer a generalisation 
of discrete convolutions \cite{Yu2016} by introducing a dilation factor $l$ 
which determines the `sampling' distance between pixels in the input. For 
$l=1$, dilated convolutions correspond to discrete convolutions.
By increasing the dilation factor, the distance between pixels sampled from the 
input becomes larger. This results in an increase in the size of the receptive 
field, without increasing the number of weights in the filter.
Figure~\ref{inp:fig:dilation} provides a visual illustration of dilated 
convolution filters.

Recent work has demonstrated that architectures using dilated convolutions are 
especially promising for image analysis tasks requiring detailed understanding 
of the scene \cite{Yu2016, Yu2017, Iizuka2017}. For inpainting the aim is to 
fill the missing region in a manner which is both globally and locally 
coherent, therefore it relies strongly on a detailed scene understanding.  In 
this work we incorporate lessons learnt from the work by Yu et al.  
\cite{Yu2016,Yu2017} and Iizuka et al. \cite{Iizuka2017} and present a 
lightweight and flexible inpainting model with minimal downsampling.

\subsection{Our contributions}

In this work we make the following four contributions.
(1) We present a light-weight and flexible inpainting model, with an order of 
magnitude fewer parameters than used in previous work.
(2) We show state-of-the-art inpainting performance on datasets of natural 
images and paintings.
(3) While acknowledging that a number of works have explored inpainting of 
cracks in paintings \cite{Giakoumis2006, Solanki2009, Cornelis2013}, we pose 
that we are the first to explore inpaintings of large regions of paintings.  
(4) We demonstrate that our model is capable of extending images (i.e., image 
extrapolation), by generating novel content which extends beyond the edges of 
the current image.

\section{Pixel Context Encoders}
\label{inp:sec:pce}

In this section we will describe our inpainting model: Pixel Context Encoders 
(PCE). Firstly, we will describe the PCE architecture, followed by details on 
the loss function used for training. 

\subsection{Architecture} 

Typically, Convolutional Neural Networks which are used for image generation 
follow an encoder-decoder type architecture \cite{Isola2016}. The encoder 
compresses the input, and the decoder uses the compressed representation (i.e., 
the bottleneck) to generate the output. Our PCE does not have a bottleneck, 
nevertheless we do distinguish between a block of layers which encodes the 
context (the encoder), and a block of layers which take the encoding of the 
image and produces the output image, with the missing region filled in (the 
decoder).

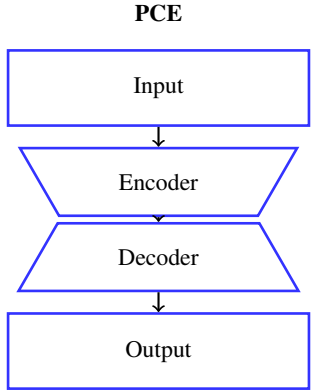
\begin{figure}[!ht]
	\centering
	\begin{tikzpicture}[every node/.style={rectangle, minimum width=1cm, 
		minimum height=1cm, line width=1pt, draw=blue!80},on grid] 
	
    \node[minimum width=4cm] (A) at (0,0) {Input};
	\node[minimum height=0.9cm, trapezium, trapezium angle=-60] (B) at 
	(0,-1.25) {Encoder};
	\node[minimum height=0.9cm, trapezium, trapezium angle=60] (C) at 
	(0,-2.25) {Decoder};
	\node[minimum width=4cm] (D) at (0,-3.5) {Output};
	\draw [->, thick] (A) -- (B);
	\draw [->, thick] (B) -- (C);
	\draw [->, thick] (C) -- (D);
	\node[above of=A, draw=none] (2) {\textbf{PCE}};

	\end{tikzpicture}
	\caption{Visualisation of the PCE architecture. Unlike traditional encoder-decoder architectures, the PCE architecture does not have a bottleneck. The encoder describes the
	context (i.e., the input), and the decoder generates the output image.}
	\label{inp:fig:nobottleneck}
\end{figure}

\textbf{The encoder} consists of two downsampling layers, followed by a block 
of $n$ dilated convolutional layers. The downsampling layers of the encoder are 
discrete convolutions with a stride of $2$. For the subsequent dilated 
convolution layers, the dilation rate $d$ increases exponentially.  The depth 
of the encoder is chosen such that the receptive field of the encoder is (at 
least) larger than the missing region, for all of our experiments $n=4$, 
resulting in a receptive field of $247 \times 247$.
Table~\ref{inp:tab:receptivefield} shows how the size of the receptive field 
grows as more layers are added to the encoder. 

By incorporating strided convolutions in the first two layers we follow Iizuka et al. \cite{Iizuka2017} and downsample the images by a factor 
$4$, our empirical results showed that this improves inpainting performance and 
drastically reduces ($5$ to $6$ times) memory requirements as compared to no 
downsampling.  We pose that the increased performance stems from the larger 
receptive field, and the local redundancy of images, i.e., neighbouring pixels 
tend to be very similar.  Nonetheless, we expect that stronger downsampling 
will results in too great of a loss of spatial resolution, lowering the 
inpainting performance.

\begin{table}
\centering
\caption{Growth of the PCE receptive field (RF) and dilation rate $d$ as a 
  function of the number of layers (depth), with a filter size of $3 \times 3$.  
The first two layers are discrete convolutions with a stride of $2$.}
  \label{inp:tab:receptivefield}
\begin{tabular}{lll}
\hline
Depth       &  $d$ & RF size \\
\hline
$1$     &   $1$     &   $3 \times 3$ \\
$2$     &   $1$     &   $7 \times 7$ \\
$3$     &   $2$     &   $23 \times 23$ \\
$4$     &   $4$     &   $55 \times 55$ \\
$5$     &   $8$     &   $119 \times 119$ \\
$6$     &   $16$     &   $247 \times 247$ \\
\hline
\end{tabular}
\end{table}

\textbf{The decoder} consists of a block of $3$ discrete convolutional layers 
which take as input the image encoding produced by the encoder. The last two 
layers of the decoder are preceded by a nearest-neighbour interpolation layer, 
which upsamples by a factor $2$, restoring the image to the original 
resolution.  Additionally, the last layer maps the image encoding back to RGB 
space (i.e., $3$ colour channels), after which all pixels which were not 
missing are restored to the ground-truth values.
 
All convolutional layers in the encoder and decoder, except for the last 
decoder layer, are followed by a Batch Normalisation layer \cite{Ioffe2015}.  
The activation functions for all convolutional layers in the encoder and 
decoder are Exponential Linear Units (ELU) \cite{Clevert2016}. 

\subsection{Loss} 

PCEs are trained through self-supervision; an image is artificially corrupted, 
and the model is trained to regress back the uncorrupted ground-truth content.  
The PCE $F$ takes an image $x$ and a binary mask $M$ (the binary mask $M$ is 
one for masked pixels, and zero for the pixels which are provided) and aims to 
generate plausible content for the masked content $F(x, M)$.  During training 
we rely on two loss functions to optimise the network: a L1 loss and a GAN 
loss.  For the GAN loss we specifically use the PatchGAN discriminator 
introduced by Isola et al. \cite{Isola2016}.

The \textbf{L1 loss} is masked such that the loss is only non-zero inside the 
corrupted region:
\begin{equation}
  \mathcal{L}_{L1} = \lVert M \odot (F(x, M) - x) \rVert_1
\end{equation}
where $\odot$ is the element-wise multiplication operation.

Generally, the \textbf{PatchGAN loss} is defined as follows:
\begin{equation}
  \min_{G} \max_{D} \mathbb{E}_{x \sim p_{data}(x)}[log(D(x) + log(1 - D(G(x))]
\end{equation}
where the discriminator $D$ aims to distinguish real from fake samples, and the 
generator $G$ aims to fool the discriminator.
For our task we adapt the loss to use our PCE as the generator:
\begin{equation}
  \begin{split}
  \mathcal{L}_{GAN} = \min_{F} \max_{D} \mathbb{E}_{x \sim 
p_{data}(x)}[log(D(x) + \\
log(1 - D(F(x, M))],
\end{split}
\end{equation}
our discriminator is similar to the global discriminator used in 
\cite{Iizuka2017}, except that we restore the ground-truth pixels before 
processing the generated image with the discriminator. This allows the 
discriminator to focus on ensuring that the generated region is consistent with 
the context.

\textbf{The overal loss} used for training thus becomes:
\begin{equation}
  \mathcal{L} = \lambda \mathcal{L}_{L1} +  (1 - \lambda) \mathcal{L}_{GAN}
\end{equation}
where $\lambda$ is fixed at $0.999$ for all experiments, following 
\cite{Pathak}.

\section{Experiments}
\label{inp:sec:exp}

To evaluate the performance of our PCE we test it on a number of datasets and 
variations of the inpainting task.  In this section we will describe the 
datasets and the experimental setting used for training, the results of image 
inpainting on $128 \times 128$ and $256 \times 256$ images, and lastly the 
image extrapolation results.

All results are reported using the Root Mean Square Error (RMSE) and Peak 
Signal Noise Ratio (PSNR) between the uncorrupted ground truth and the output 
produced by the models.

\subsection{Datasets}

\textbf{ImageNet.} As a set of natural images we use the subset of $100,000$ 
images that Pathak et al.  \cite{Pathak} selected from the ImageNet dataset 
\cite{Russakovsky2014}. The performance is reported on the complete ImageNet 
validation set consisting of $50,000$ images.

\textbf{PaintersN.}
The ``\textit{Painters by Numbers}'' dataset (PaintersN) as published on 
Kaggle\footnote{\url{https://www.kaggle.com/c/painter-by-numbers}} consists of 
$103,250$ photographic reproductions of artworks by well over a thousand 
different artists.  The dataset is divided into a training set ($93,250$ 
images), validation set ($5000$ images), and test set ($5000$ images) used for 
training the model, optimising hyper-parameters, and reporting performances, 
respectively. 

For both datasets all images were scaled such that the shortest side was $256$ 
pixels, and then a randomly located $256 \times 256$ crop was extracted.

\subsection{Experimental settings}

In this section the details on the settings of the hyperparameters and training 
procedure are provided. The layers of the encoder and the decoder consist of 
$128$ filters with spatial dimensions of $3 \times 3$ for all experiments in 
this work.  All dilated layers were initialised using identity initialisation 
cf.  \cite{Yu2016}, which sets the weights in the filter such that initially 
each layer simply passes its input to the next.  All discrete convolutional 
layers were initialised using Xavier initialisation \cite{Glorot2010}.

The PatchGAN discriminator we used consists of $5$ layers of filters with 
spatial dimensions of $3 \times 3$, using LeakyReLU as the activation function 
($\alpha = 0.2$).  For the first $4$ layers the number of filters increases 
exponentially (i.e., $64$, $128$, $256$, $512$), the 5th layer outputs a single 
channel, the real/fake judgement for each patch in the input.

The network was optimised using ADAM \cite{Kingma2015} until the L1 loss on the 
training set stopped decreasing. We were able to consider the training loss as 
we found that there was no real risk of overfitting. Probably, this is due to  
the low number of model parameters. The size of the minibatches varied 
depending on memory capabilities of the graphics card.

All images were scaled to the target resolution using bilinear interpolation 
when necessary. During training the data was augmented by randomly horizontally 
flipping the images. 

Using the hyperparameter settings specified above, our PCE model has 
significantly fewer model parameters than previously presented inpainting 
models.  Table~\ref{inp:tab:modelparam} gives an overview of the model 
parameters\footnote{At the time of writing the exact implementation by Iizuka 
et al. was not available, therefore we calculated the number of parameters 
based on the sizes of the weight matrices given in \cite{Iizuka2017}, thus not counting 
any bias, normalisation, or additional parameters.} for the most relevant 
models. Clearly, the number of parameters of the PCE model is much smaller than 
those of comparable methods.

\begin{table}
\centering
\caption{Number of parameters for the generators and discriminators of the 
inpainting models by Pathak et al.  \cite{Pathak}, Iizuka et al.  
\cite{Iizuka2017}, and ours.}
  \label{inp:tab:modelparam}
\begin{tabular}{lrr}
  \hline
 Model & \# Generator &  \# Discriminator \\
\hline
CE \cite{Pathak} &  $71,130,531$ & $2,766,529$ \\
Iizuka et al. \cite{Iizuka2017} & $6,061,600$ & $29,322,624$ \\
PCE &  $1,041,152$ & $1,556,416$ \\
\hline
\end{tabular}
\end{table}

\subsection{Region inpainting}
A commonly performed task to evaluate inpainting, is region inpainting 
\cite{Pathak, Yeh2016, yang2017}. Typically, in region inpainting a quarter of 
all the pixels are removed by masking the centre of the image (i.e., centre 
region inpainting).  This means that for a $256 \times 256$ image the central 
$128 \times 128$ region is removed. A variant of centre region inpainting is 
random region inpainting where the missing region is not fixed to the centre of 
the image, but is placed randomly.  This requires the model to learn to inpaint 
the region independently of where the region is, forcing it to be more 
flexible. 

In this section we will first present results of centre region image inpainting 
on $128 \times 128$ images, followed by the results of centre and random region 
inpainting on $256 \times 256$ images.

To evaluate the centre-region inpainting performance of our PCE model we 
compare it against the performance of the CE model by Pathak et al.  
\cite{Pathak}.  For this reason, we initially adopt the maximum resolution of 
the model of Pathak et al., i.e., $128 \times 128$, subsequent results will be 
presented on $256 \times 256$ images. For the $128 \times 128$ ImageNet 
experiments we use the pretrained model release by Pathak et al. For the $128 
\times 128$ results on the PaintersN dataset we have trained their model from 
scratch. 

The model by Pathak et al.  \cite{Pathak} uses an overlap (of $4$ pixels) 
between the context and the missing region.  Their intention with this overlap 
is to improve consistency with the context, but as a consequence it also makes 
the task slightly easier, given that the masked region shrinks by $4$ pixels on 
all sides.  For all centre region inpainting experiments we also\footnote{PCE 
do not require this overlap to achieve a smooth transition between the context 
and the missing region. Nonetheless we incorporate to make it a fair 
comparison.} add a $4$ pixel overlap between the context and the missing 
region, however unlike Pathak et al.  we do not use a higher weight for the 
loss in the overlapping region, as our model is able to achieve local 
consistency without this additional encouragement.

In Table~\ref{inp:tab:cr_128_results} the results on $128 \times 128$ images 
are shown, all models are trained and evaluated on both the ImageNet and 
dataset the PaintersN dataset, to explore the generalisability of the models.  
The performance of our PCE model exceeds that of the model by Pathak for both 
datasets. Nonetheless, both models perform better on the PaintersN dataset, 
implying that this might be an easier dataset to inpaint on. Overall, our PCE 
model trained on the $100,000$ image subset of ImageNet performs best, 
achieving the lowest RMSE and highest PSNR on both datasets.  

\begin{table}
\footnotesize
\centering
\caption{Center region inpainting results on $128 \times 128$ images with a $64 
  \times 64$ masked region. RMSE and PSNR for models trained on the ImageNet 
and PaintersN datasets (horizontally), and evaluated on both datasets 
(vertically).}
  \label{inp:tab:cr_128_results}
\begin{tabular}{llll|ll}
  \hline
& &  \multicolumn{2}{c|}{ImageNet} & \multicolumn{2}{c}{PaintersN} \\
\hline
Trained on & Model & RMSE & PSNR & RMSE & PSNR \\
\hline
\multirow{2}{*}{Imagenet} & CE \cite{Pathak} &  $43.12$ & $15.44$ & $40.69$ & 
$15.94$  \\
& PCE &  $\mathbf{22.88}$ & $\mathbf{20.94}$  & $\mathbf{22.53}$ & 
$\mathbf{21.08}$ \\
\hline
\multirow{2}{*}{PaintersN} & CE \cite{Pathak} &  $43.69$ & $15.32$ & $40.58$ & 
$15.96$  \\
& PCE &  $24.35$ & $20.40$ & $23.33$ & $20.77$  \\
\hline
\end{tabular}
\end{table}

\begin{figure*}[!t]
\begin{tabular}{cccc}
  \footnotesize{Ground Truth} & \footnotesize{Input} & CE & PCE \\
\includegraphics[width=.24\linewidth]{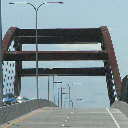} 
&
\includegraphics[width=.24\linewidth]{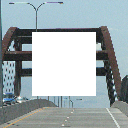} &
\includegraphics[width=.24\linewidth]{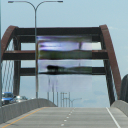} &
\includegraphics[width=.24\linewidth]{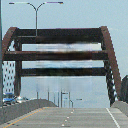} \\
\includegraphics[width=.24\linewidth]{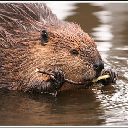} &
\includegraphics[width=.24\linewidth]{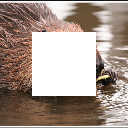} &
\includegraphics[width=.24\linewidth]{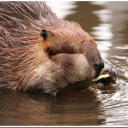} &
\includegraphics[width=.24\linewidth]{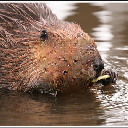} \\
\includegraphics[width=.24\linewidth]{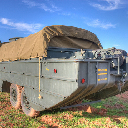} &
\includegraphics[width=.24\linewidth]{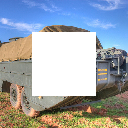} &
\includegraphics[width=.24\linewidth]{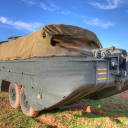} &
\includegraphics[width=.24\linewidth]{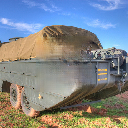} \\
\includegraphics[width=.24\linewidth]{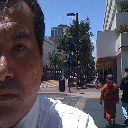} &
\includegraphics[width=.24\linewidth]{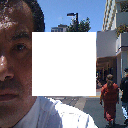} &
\includegraphics[width=.24\linewidth]{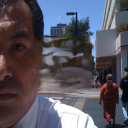} &
\includegraphics[width=.24\linewidth]{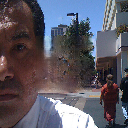} \\
  \end{tabular}
  \caption{Comparison between CE \cite{Pathak} and our PCE model, on inpainting 
  a $64 \times 64$ region in $128 \times 128$ images taken from the ImageNet 
validation set.}
  \label{inp:fig:128_examples_im}
\end{figure*}

\begin{figure*}[!t]
\begin{tabular}{cccc}
  \footnotesize{Ground Truth} & \footnotesize{Input} & CE & PCE \\
\includegraphics[width=.24\linewidth]{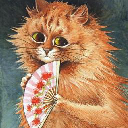} 
&
\includegraphics[width=.24\linewidth]{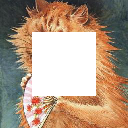} &
\includegraphics[width=.24\linewidth]{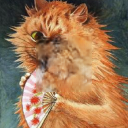} &
\includegraphics[width=.24\linewidth]{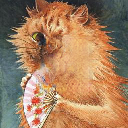} \\
\includegraphics[width=.24\linewidth]{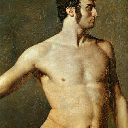} &
\includegraphics[width=.24\linewidth]{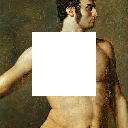} &
\includegraphics[width=.24\linewidth]{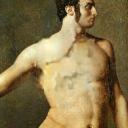} &
\includegraphics[width=.24\linewidth]{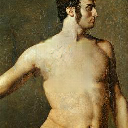} \\
\includegraphics[width=.24\linewidth]{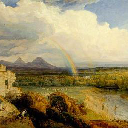} &
\includegraphics[width=.24\linewidth]{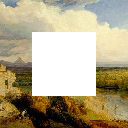} &
\includegraphics[width=.24\linewidth]{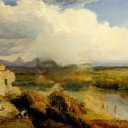} &
\includegraphics[width=.24\linewidth]{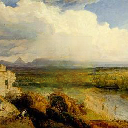} \\
\includegraphics[width=.24\linewidth]{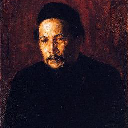} &
\includegraphics[width=.24\linewidth]{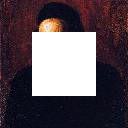} &
\includegraphics[width=.24\linewidth]{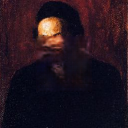} &
\includegraphics[width=.24\linewidth]{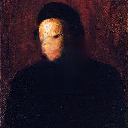} \\
  \end{tabular}
  \caption{Comparison between CE \cite{Pathak} and our PCE model, on inpainting 
  a $64 \times 64$ region in $128 \times 128$ images taken from the PaintersN 
test set.}
  \label{inp:fig:128_examples_wi}
\end{figure*}

Additionally, in Figures~\ref{inp:fig:128_examples_im} and 
\ref{inp:fig:128_examples_wi} we show examples of centre region inpainting on 
the ImageNet and PaintersN datasets, respectively.
Qualitatively, our PCE model appears to generate content which is less blurry
and more consistent with the context. Obviously, both models struggle to 
recover content which was not available in the input, but when not considering 
the ground truth, and only the generated output, we observe that our PCE model 
produces more plausible images.

As our PCE model is capable of inpainting images larger than $128 \times 128$, 
we show results on $256 \times 256$ images, with a $128 \times 128$ missing 
region in Table~\ref{inp:tab:cr_256_results}. Additionally, in this table we 
also show random region inpainting results.  The random region inpainting 
models were trained without overlap between the context and the missing region. 

\begin{table}
\footnotesize
\centering
\caption{Center and random region inpainting results for PCE on $256 \times 
  256$ images with a $128 \times 128$ masked region. RMSE and PSNR for models 
trained on the ImageNet and PaintersN datasets (horizontally), and evaluated on 
both datasets (vertically). The first two rows are for centre region 
inpainting, the last two for random region inpainting.}
  \label{inp:tab:cr_256_results}
\begin{tabular}{llll|ll}
  \hline
& &  \multicolumn{2}{c|}{ImageNet} & \multicolumn{2}{c}{PaintersN} \\
\hline
Region & Trained on &  RMSE & PSNR & RMSE & PSNR \\
\hline
\multirow{2}{*}{Center} & Imagenet & $24.36$ & $20.40$ & $23.87$ & $20.57$ \\
& PaintersN &  $24.99$ & $20.17$ & $23.41$  & $20.74$  \\
\hline
\multirow{2}{*}{Random} & Imagenet &  $24.62$ & $20.30$  & $24.20$ & $20.45$ \\
& PaintersN & $25.13$ & $20.13$ & $24.06$ & $20.51$  \\
\hline
\end{tabular}
\end{table}

The results in Table~\ref{inp:tab:cr_256_results} not only show that our model 
is capable of inpainting images at a higher resolution, they also show that 
randomising the location of the missing region only has a minimal effect on the 
performance of our model. Although all results were obtained by training a 
model specifically for the task, we note that no changes in model configuration 
were necessary to vary between tasks.  

\begin{figure}[!t]
\begin{tabular}{ccc}
  \footnotesize{Ground Truth} & \footnotesize{Input} & PCE \\
\includegraphics[width=.29\linewidth]{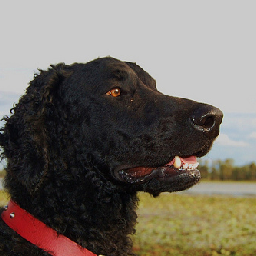} 
&
\includegraphics[width=.29\linewidth]{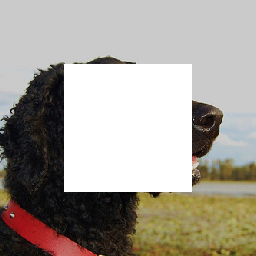} &
\includegraphics[width=.29\linewidth]{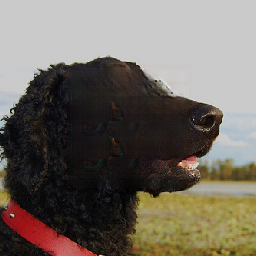} \\
\includegraphics[width=.29\linewidth]{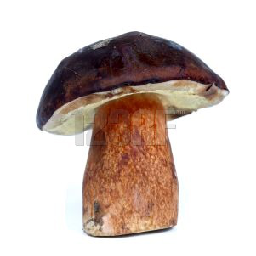} &
\includegraphics[width=.29\linewidth]{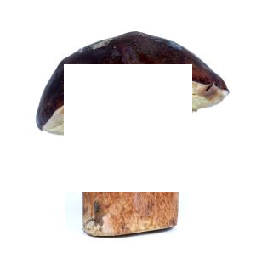} &
\includegraphics[width=.29\linewidth]{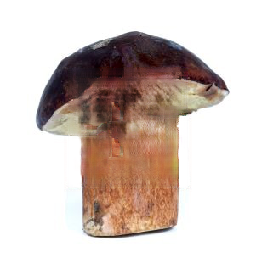} \\

\includegraphics[width=.29\linewidth]{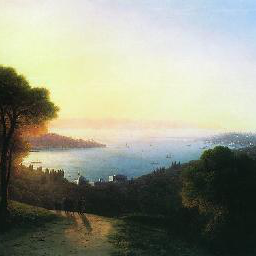} 
&
\includegraphics[width=.29\linewidth]{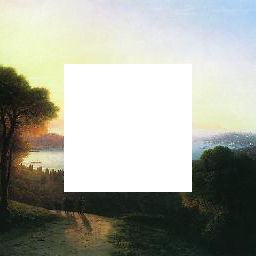} &
\includegraphics[width=.29\linewidth]{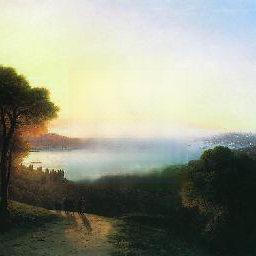} \\
\includegraphics[width=.29\linewidth]{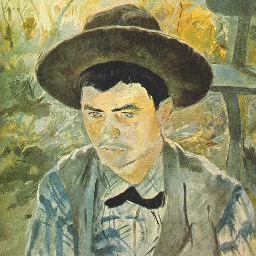} &
\includegraphics[width=.29\linewidth]{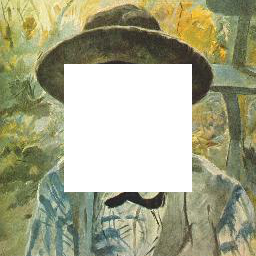} &
\includegraphics[width=.29\linewidth]{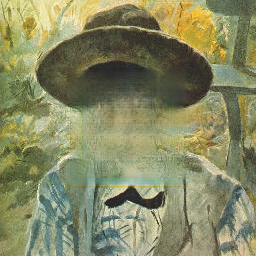} \\
  \end{tabular}
  \caption{Examples produced by our PCE model, on inpainting a $128 \times 128$ 
  region in $256 \times 256$ images taken from the ImageNet validation set in 
the first two rows, and PaintersN test set in the last two rows.}
  \label{inp:fig:256_examples}
\end{figure}

In Figure~\ref{inp:fig:256_examples} we show several centre region inpainting 
examples generated by our PCE model on $256 \times 256$ images.

\subsection{Image extrapolation}

In this section, we explore image extrapolation; generating novel content 
beyond the image boundaries. By training a PCE to reconstruct the content on 
the boundary of an image (effectively inverting the centre region mask), we are 
able to teach the model to extrapolate images. In 
Table~\ref{inp:tab:extra_results} we show the results of image extrapolation 
obtained by only providing the centre $192 \times 192$ region of $256 \times 
256$ images, aiming to restore the $64$ pixel band surrounding it.  For our 
region inpainting experiments we corrupted $\frac{1}{4}$th of the pixels, 
whereas for this task $\frac{9}{16}$th of the pixels are corrupted. Despite the 
increase in size of the reconstructed region, the difference in performance is 
not very large, highlighting the viability of image extrapolation with this 
approach.

\begin{table}
\footnotesize
\centering
\caption{Image extrapolation results for PCE on $256 \times 256$ images based 
  on a provided $192 \times 192$ centre region. RMSE and PSNR for models 
trained on the ImageNet and PaintersN datasets (horizontally), and evaluated on 
both datasets (vertically).}
  \label{inp:tab:extra_results}
\begin{tabular}{lll|ll}
  \hline
 &  \multicolumn{2}{c|}{ImageNet} & \multicolumn{2}{c}{PaintersN} \\
\hline
Trained on &  RMSE & PSNR & RMSE & PSNR \\
\hline
Imagenet & $31.81$ & $18.08$ & $32.82$ & $17.81$ \\
PaintersN & $32.67$ & $17.85$ & $32.39$ & $17.92$  \\
\hline
\end{tabular}
\end{table}

\begin{figure}[!t]
\begin{tabular}{ccc}
  \footnotesize{Ground Truth} & \footnotesize{Input} & PCE \\
\includegraphics[width=.29\linewidth]{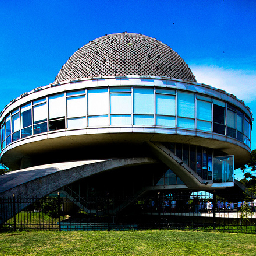} 
&
\includegraphics[width=.29\linewidth]{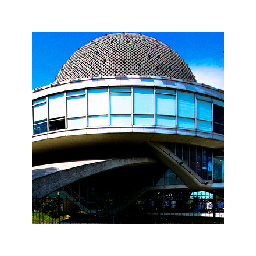} 
&
\includegraphics[width=.29\linewidth]{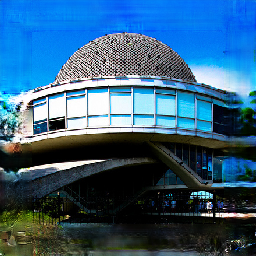} 
\\
\includegraphics[width=.29\linewidth]{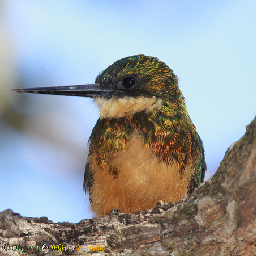} 
&
\includegraphics[width=.29\linewidth]{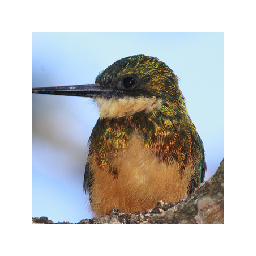} 
&
\includegraphics[width=.29\linewidth]{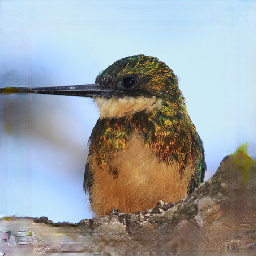} 
\\

\includegraphics[width=.29\linewidth]{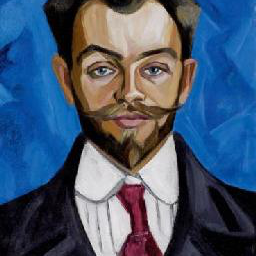} 
&
\includegraphics[width=.29\linewidth]{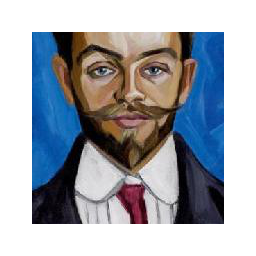} 
&
\includegraphics[width=.29\linewidth]{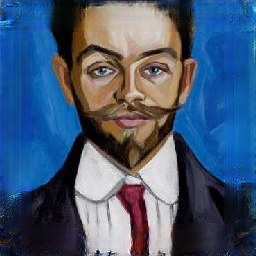} 
\\
\includegraphics[width=.29\linewidth]{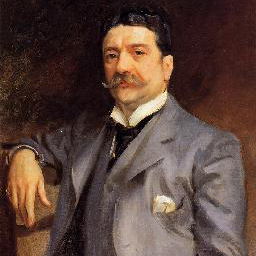} 
&
\includegraphics[width=.29\linewidth]{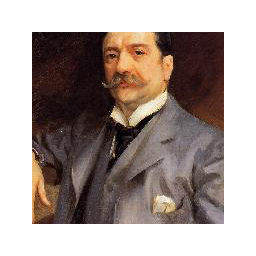} 
&
\includegraphics[width=.29\linewidth]{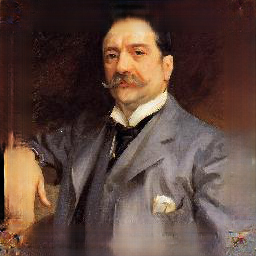} 
  \\
  \end{tabular}
  \caption{Examples produced by our PCE model, on extrapolating $192 \times 
  192$ regions taken from $256 \times 256$ images from the ImageNet validation 
set in the first two rows, and PaintersN test set in the last two rows.}
  \label{inp:fig:extrapolate}
\end{figure}

In Figure~\ref{inp:fig:extrapolate} we show four examples obtained through 
image extrapolation. Based on only the provided input our PCE is able to 
generate novel content for the $64$ pixel band surrounding the input.  Although 
the output does not exactly match the input, the generated output does appear 
plausible. 

\begin{figure*}[!t]
\begin{tabular}{cccc}
  \footnotesize{Original} & PCE &
  \footnotesize{Original} & PCE \\
\includegraphics[width=.24\linewidth,cfbox=black!20 0.5pt 
0pt]{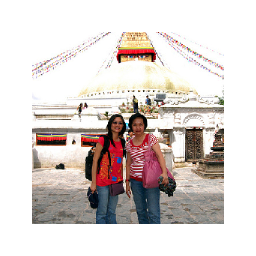} &
\includegraphics[width=.24\linewidth,cfbox=black!20 0.5pt 0pt]{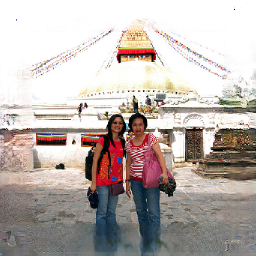} &
\includegraphics[width=.24\linewidth,cfbox=black!20 0.5pt 0pt]{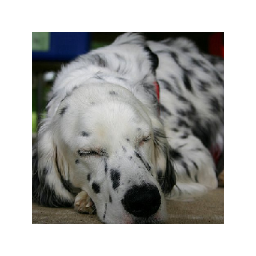} &
\includegraphics[width=.24\linewidth,cfbox=black!20 0.5pt 0pt]{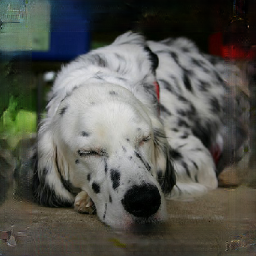} \\
\includegraphics[width=.24\linewidth,cfbox=black!20 0.5pt 0pt]{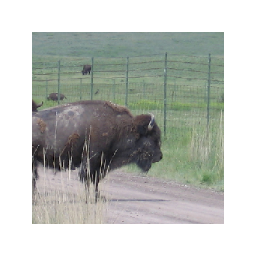} &
\includegraphics[width=.24\linewidth,cfbox=black!20 0.5pt 0pt]{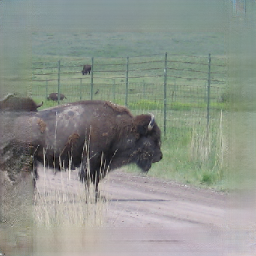} &
\includegraphics[width=.24\linewidth,cfbox=black!20 0.5pt 0pt]{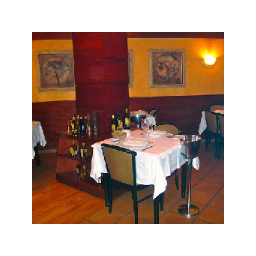} &
\includegraphics[width=.24\linewidth,cfbox=black!20 0.5pt 0pt]{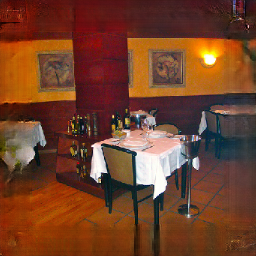} \\

\includegraphics[width=.24\linewidth,cfbox=black!20 0.5pt 
0pt]{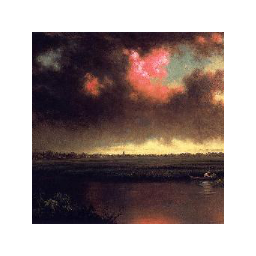} &
\includegraphics[width=.24\linewidth,cfbox=black!20 0.5pt 0pt]{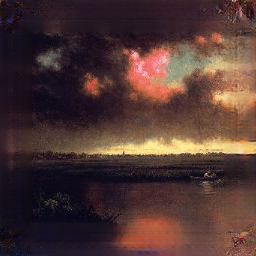} &
\includegraphics[width=.24\linewidth,cfbox=black!20 0.5pt 0pt]{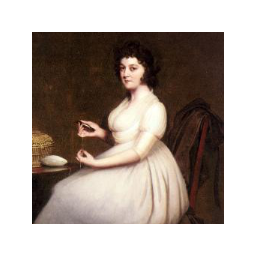} &
\includegraphics[width=.24\linewidth,cfbox=black!20 0.5pt 0pt]{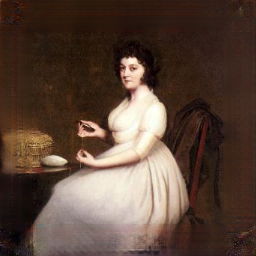} \\
\includegraphics[width=.24\linewidth,cfbox=black!20 0.5pt 0pt]{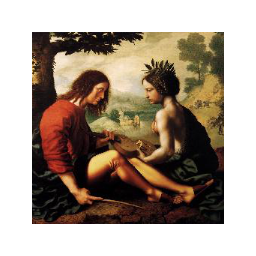} &
\includegraphics[width=.24\linewidth,cfbox=black!20 0.5pt 0pt]{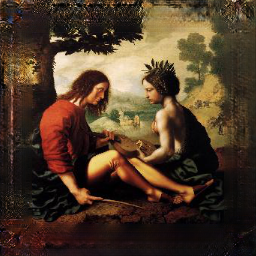} &
\includegraphics[width=.24\linewidth,cfbox=black!20 0.5pt 0pt]{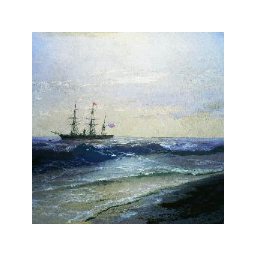} &
\includegraphics[width=.24\linewidth,cfbox=black!20 0.5pt 0pt]{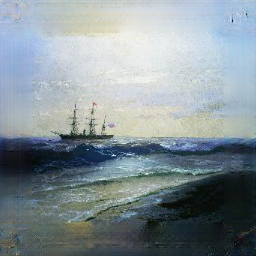} \\
  \end{tabular}
  \caption{Examples produced by our PCE model, on extrapolating $192 \times 
  192$ images beyond their current boundaries. The images in the first two rows 
are from the ImageNet validation, and in the last two rows from the PaintersN 
test set.}
  \label{inp:fig:extend}
\end{figure*}

Additionally, in Figure~\ref{inp:fig:extend} we show images obtained by 
applying the PCE trained for image extrapolation to uncorrupted images, resized 
to $192 \times 192$ pixels.  By resizing the images to the resolution of the 
region the model was train on, the model will generate a band of $64$ pixels of 
novel content, for which there is no ground truth.

\section{Conclusion}
\label{inp:sec:conclusion}

In this paper we presented a novel inpainting model: Pixel Content Encoders 
(PCE), by incorporating dilated convolutions and PatchGAN we were able to 
reduce the complexity of the model as compared to previous work. Moreover, by 
incorporating dilated convolutions PCE are able to preserve the spatial 
resolution of images, as compared to encoder-decoder
style architectures which lose spatial information by compressing the input into `bottleneck' representations. 

We trained and evaluated the inpainting performance of PCE on two datasets of natural images
and paintings, respectively. The results show that regardless of the dataset PCE
were trained on they outperform previous work on either dataset, even when considering
cross-dataset performance (i.e., training on natural images and evaluating on paintings, 
and vice versa). Based on the cross-dataset performance we pose that PCE solve the inpainting problem in a largely data-agnostic manner. By encoding the context surrounding the missing region PCE are able to generate plausible content for the missing region in a manner that is coherent with the context.

The approach presented in this paper does not explicitly take into account the 
artist's style.  However, we would argue that the context reflects the artist's 
style, and that generated content coherent with the context is therefore also 
reflects the artist's style. Future research on explicitly incorporating the 
artist's style is necessary to determine whether it is beneficial for 
inpainting on artworks to encode the artist's style in addition to the context. 

We conclude that PCE offer a promising avenue for image inpainting and image extrapolation. 
With an order of magnitude fewer model parameters than previous inpainting models, PCE obtain state-of-the-art performance on benchmark datasets of natural images and paintings.  
Moreover,  due to the flexibility of the PCE architecture it can be 
used for other image generation tasks, such as image extrapolation.  We 
demonstrate the image extrapolation capabilities of our model by restoring 
boundary content of images, and by generating novel content beyond the existing 
boundaries.

% \section{Acknowledgements}
% TODO don't forget this!

{\small
\bibliographystyle{ACM-Reference-Format}
\bibliography{library.bib}
}

\end{document}